\documentclass[final]{cvpr}

\usepackage{url}            
\usepackage{booktabs}       
\usepackage{amsmath,bm}
\usepackage{graphicx}  
\usepackage{ntheorem}
\usepackage{wrapfig}
\usepackage{subfigure}
\usepackage{times}
\usepackage{epsfig}
\usepackage{amsmath}
\usepackage{amssymb}
\usepackage{graphicx}
\usepackage[T1]{fontenc}
\usepackage{threeparttable}

\usepackage[font=small,skip=3pt]{caption}
\graphicspath{{./}}

\def\cvprPaperID{944} 
\def\confYear{CVPR 2021}

\begin{document}

\title{Deep Implicit Templates for 3D Shape Representation}

\author{%
  Zerong Zheng \qquad  Tao Yu  \qquad  Qionghai Dai \qquad  Yebin Liu \\
  \qquad  \\ 
  Department of Automation, Tsinghua University, Beijing, China
}

\maketitle


\pagestyle{empty}  
\thispagestyle{empty} 

\begin{abstract}
Deep implicit functions (DIFs), as a kind of 3D shape representation, are becoming more and more popular in the 3D vision community due to their compactness and strong representation power. However, unlike polygon mesh-based templates, it remains a challenge to reason dense correspondences or other semantic relationships across shapes represented by DIFs, which limits its applications in texture transfer, shape analysis and so on. To overcome this limitation and also make DIFs more interpretable, we propose Deep Implicit Templates, a new 3D shape representation that supports explicit correspondence reasoning in deep implicit representations. Our key idea is to formulate DIFs as \emph{conditional deformations of a template implicit function}. 
To this end, we propose Spatial Warping LSTM, which decomposes the conditional spatial transformation into multiple point-wise transformations and guarantees generalization capability. 
Moreover, the training loss is carefully designed in order to achieve high reconstruction accuracy while learning a plausible template with accurate correspondences in an unsupervised manner.
Experiments show that our method can not only learn a common implicit template for a collection of shapes, but also establish dense correspondences across all the shapes simultaneously without any supervision. 

\end{abstract}

\section{Introduction}
\label{sec:intro}

\begin{figure*}
	\centering
	\includegraphics[width=0.30\linewidth]{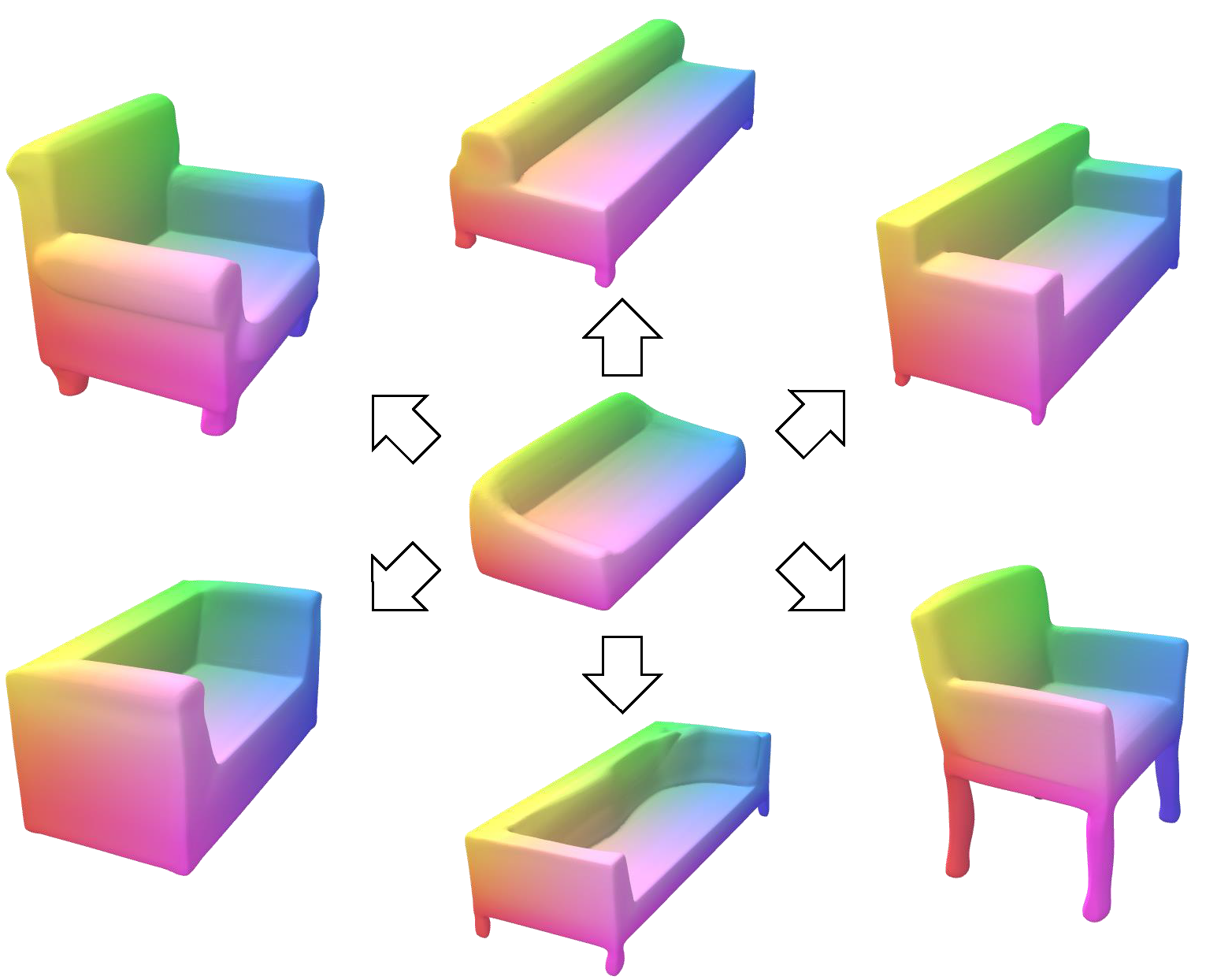} \hfill
	\includegraphics[width=0.30\linewidth]{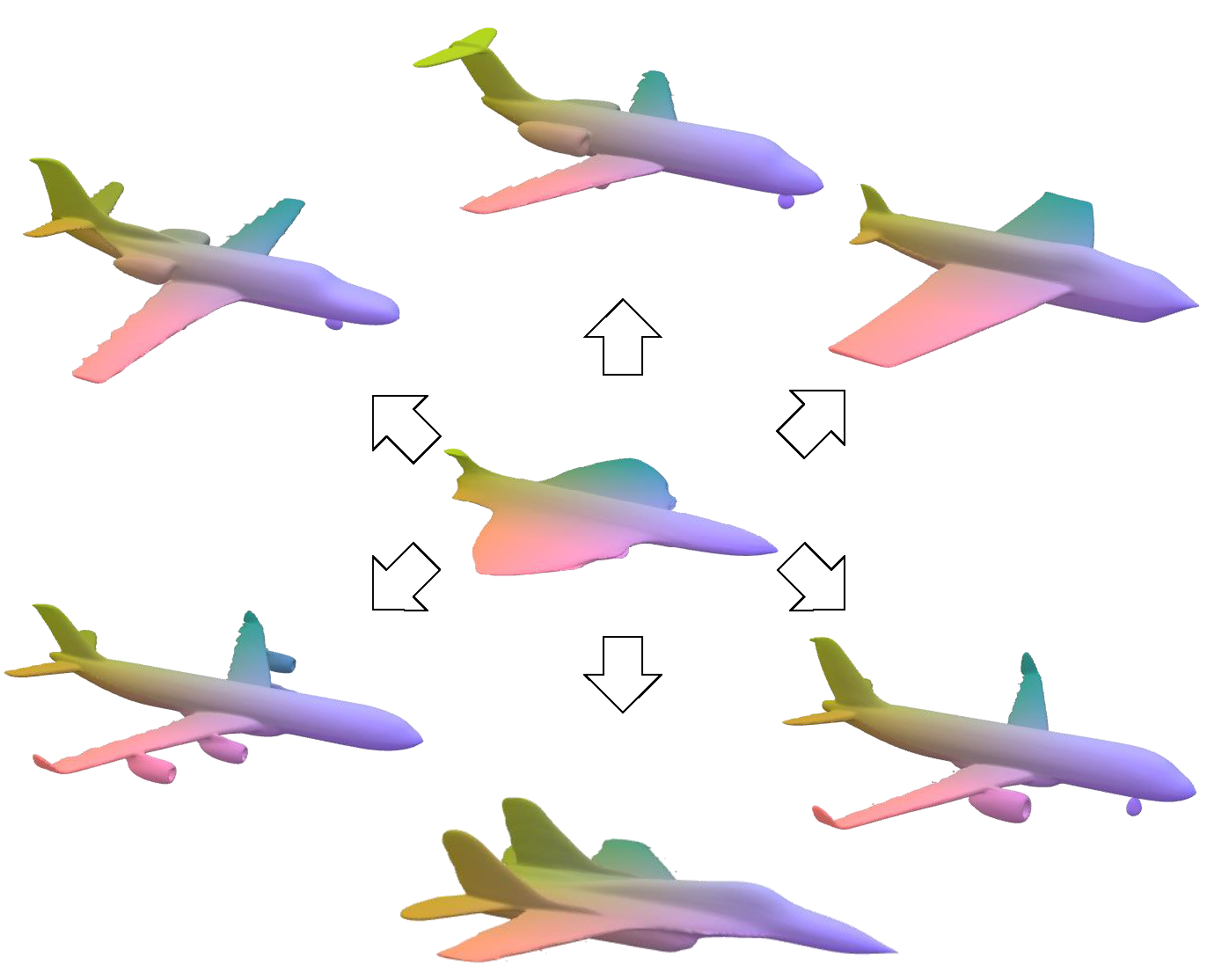} \hfill
    \includegraphics[width=0.30\linewidth]{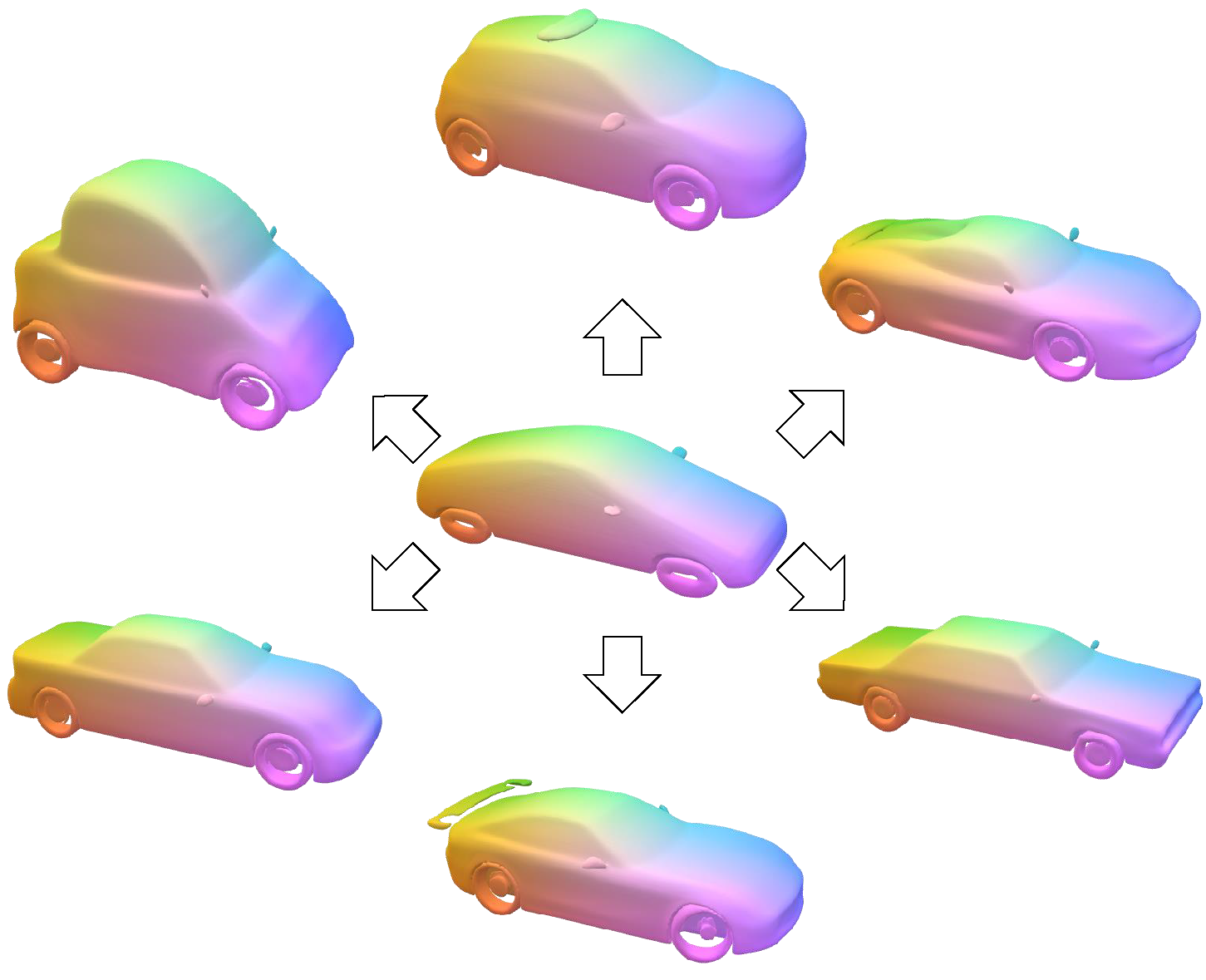} \hfill
    \caption{Example results of our representation. Our approach is able to factor out a plausible template (middle) from a set of shapes (surroundings), and builds up dense correspondences (color-coded) across all the shapes automatically without any supervision. }
	\label{fig:teaser}
\end{figure*}

Representing 3D objects effectively and efficiently in neural networks is fundamental for many tasks in computer vision, including 3D model reconstruction, matching, manipulation and understanding.  
In the pioneering studies, researchers have adopted various traditional geometry representations, including voxel grids \cite{3dgan,3dShapeNets,MarrNet2017,shapehd,HierarchicalSurface,OctreeGeneratingNetworks}, point clouds \cite{zamorski2018adversarial,achlioptas2017latent_pc,FoldingNet2018,PUGAN} and meshes \cite{HMD2019,tex2shape2019,Pixel2Mesh2018,Pixel2MeshPlusPlus,AtlasNet2018,AtlasNetV2,multichart}.  
In the past several years, deep implicit functions (DIFs) have been proposed as an alternative \cite{park2019deepsdf,Mescheder2019OccupancyNetwork,chen2019implicit,DSIF,implicit_geometric_reg,DISN,local_implicit_grid,deep_local_shapes,zekun2020dualsdf,Tretschk2020PatchNets,duan2020cdeepsdf}. 
Compared to traditional representations, DIFs show expressive and flexible capacity for representing complex shapes and fine geometric details, even in challenging tasks like human digitization \cite{pifuSHNMKL19}. 

Unfortunately, the implicit nature of DIFs is also its Achilles' Heels: 
although DIFs are good at approximating individual shapes, they provide no information about the relationship between two different ones. One can easily establish vertice-to-vertice correspondences between two shapes when using mesh templates \cite{3DCODED,Pixel2Mesh2018,AtlasNetV2}, but that is difficult in DIFs. The lack of semantic relationship in DIFs poses significant challenges for using DIFs in downstream applications such as shape understanding and editing. 

To overcome this limitation, we propose \textit{Deep Implicit Templates}, a new way to interpret and implement DIFs. The key idea is to decompose a conditional deep implicit function into two components: a template implicit function and a conditional spatial warping function. The template implicit function represents the ``mean shape'' for a category of objects, while the spatial warping function deforms the template implicit function to form specific object instances. 
On one hand, as both the template and the warping field are defined in an implicit manner, the advantages of deep implicit representations (compactness and efficiency) are preserved. On the other hand, with the template implicit function as an intermediate shape, the warping function automatically establishes dense correspondences across different object. 

More recently, some techniques use a set of primitives to represent 3D shapes in order to capture structure-level semantics \cite{SIF,DSIF,AtlasNet2018,AtlasNetV2}. 
The primitives can be either manually defined \cite{SIF,AtlasNet2018} or learned from data \cite{DSIF,AtlasNetV2}. 
We emphasize that our method is essentially different from them in two ways. 
First, our method decomposes the implicit representations into a template implicit function and a continuous warping field. Compared to element-based methods, our decomposition not only provides a complete, global template for the training data, enabling many applications such as uv mapping and keypiont labeling, but also makes the latent shape space more interpretable as we can inspect how shapes deform. 
Second, our method directly builds up accurate correspondences in the whole 3D space, while element-based methods rely on interpolation or feature matching to compute dense correspondences. Overall, our method provides more flexibility and scalability to control the template and/or its deformation: one can, for example, replace the template in our framework with a custom designed one without losing the representation power, or apply additional semantic constraints on the spatial deformation for specific problems such as dynamic human modeling (Sec.\ref{sec:extension}).

Training deep implicit templates is not straight-forward, because we have no access to either the ground-truth mapping between the templates and shape instances, or dense correspondence annotations across different shapes.
Our ultimate goal is to make deep implicit templates an effective representation that can: 1) represent training shapes accurately, 2) establish plausible correspondences across shapes and 3) generalize to unseen data. 
However, without proper design and regularization, the network may not be able to learn such a representation in an unsupervised manner. 
We make several technical contributions to address these challenges. 
In terms of network architecture, we propose \emph{Spatial Warping LSTM}, which decomposes the conditional spatial warping into multi-step point-wose transformation, guaranteeing the generalization capacity and the representation power of our warping function. 
In addition, we introduce \emph{a progressive reconstruction loss} for our Spatial Warping LSTM, which further improves the reconstruction accuracy.  \emph{Two-level regularization} is also proposed to obtain plausible templates with accurate correspondences in an unsupervised manner.   
As shown in the experiments, our method can learn a plausible implicit template for a set of shapes, with conditional warping fields that accurately represent shapes while establishing dense correspondences among them without any supervision (See Fig.\ref{fig:teaser}). 
Overall, the proposed Deep Implicit Templates significantly expands the capability of DIFs without losing its advantages, making it a more effective implicit representation for 3D learning tasks. Code is available at \url{https://github.com/ZhengZerong/DeepImplicitTemplates}.



\section{Related Work}
\label{sec:related_work}

\noindent\textbf{Deep Implicit Functions (DIFs).}
Implicit functions represent shapes by constructing a continuous volumetric field and embedding shapes as its iso-surface \cite{implicit_rbf_representation,implicit_polygon_soup,implicit_surface_interpolate}. In recent years, implicit functions have been introduced into neural networks  \cite{park2019deepsdf,Mescheder2019OccupancyNetwork,chen2019implicit,implicit_geometric_reg,DISN,pifuSHNMKL19,local_implicit_grid,deep_local_shapes,zekun2020dualsdf,Tretschk2020PatchNets,duan2020cdeepsdf} and show promising results. For example, DeepSDF \cite{park2019deepsdf} proposed to learn an implicit function where the network output represents the signed distance of the point to its nearest surface. Other approaches defined the implicit functions as 3D occupancy probability and turned shape representation into a point classification problem \cite{Mescheder2019OccupancyNetwork,chen2019implicit,DISN}. Some latest studies proposed to blend multiple local implicit functions in order to generalize to more complex scenes as well as to capture more geometric details \cite{local_implicit_grid,deep_local_shapes,Tretschk2020PatchNets}. 
The training loss used in DeepSDF is also improved for more accurate reconstruction \cite{duan2020cdeepsdf}. 
DualSDF \cite{zekun2020dualsdf} extended DeepSDF by introducing a coarse layer to support shape manipulation.  Occupancy Flow \cite{OccupancyFlow}, from another aspect, extended 3D occupancy functions into 4D domains, but this method is restricted to represent temporally continuous 4D sequences. 
Overall, implicit functions are promising for representing complex shapes and detailed surfaces, but it remains difficult to reason dense correspondences between different shapes represented by DIFs. 
In contrast, our method and other concurrent works~\cite{Liu2020LIFC,deng2020DIFNet} overcomes this limitation and expands the capability of DIFs by introducing dense correspondences across shapes into DIFs. 

\noindent\textbf{Elementary Structures.} 
Elementary representations, which aim to describe complex shapes uing a collection of simple shape elements, have been extensively studied for many years in computer vision and graphics \cite{Survey_Simple_Geometric_Primitives,GlobFit2011,CompletionAndReconstruction,JointShapeSeg}. In this direction, previous methods usually require complicated non-convex optimization for primitive fitting. 
In order to improve the efficiency and effectiveness, various deep learning techniques were adopted, such as recurrent networks \cite{3DPRNN}, differential model estimation \cite{supervised_fitting} and unsupervised training losses \cite{Assembling_Volumetric_Primitives}. 
Some recent approaches introduced more complex shape elements, such as multiple charts \cite{multichart}, part segmentations \cite{GRASS2017,Meta2014,Paschalidou2020CVPR}, axis-aligned 3D Gaussians \cite{SIF}, local deep implicit functions \cite{DSIF}, superquadrics \cite{Superquadrics}, convex decomposition \cite{CvxNet}, box-homeomorphic parts \cite{gaosdmnet2019} or learnable shape primitives \cite{AtlasNetV2,partbased_template,Assembling_Volumetric_Primitives,Coabstraction2012}. Although elementary representation is compact, it is challenging to produce consistent fitting across different shapes. 
In addition, local elements cannot be used in many tasks (e.g., shape completion) due to the lack of global knowledge.

\noindent\textbf{Template Learning.} 
Fitting global shape templates instead of local primitives has also attracted a lot of research efforts, as mesh-based templates are able to efficiently represent similar shapes such as articulated human bodies \cite{SMPL2015,SCAPE2005,HumanBodySpace,StitchedPuppet}. They are very popular in many data-driven 3D reconstruction studies. 
For example, articulated human body templates, such as SMPL \cite{SMPL2015}, are now widely used in a lot of human modeling studies \cite{hmr,kolotouros2019cmr,HMD2019,tex2shape2019}. Ellipsoid meshes as a much simpler template can also be deformed to represent various objects \cite{Pixel2Mesh2018,Pixel2MeshPlusPlus}. Given a predefined template, some recent techniques like 3D-Coded \cite{3DCODED} learned to perform shape matching in unsupervised manner. 
Although using mesh-based templates is convenient, mesh-based templates are unable to deal with topological changes and require dense vertices to recover surface details \cite{Pixel2Mesh2018}.  Moreover, when representing shapes that are highly different from the template,  mesh-based templates suffer from deformation artifacts (e.g., triangle intersection, extreme distortion, etc.) due to large number of degrees of freedom for mesh vertex coordinates.
In contrast, the template in our method is defined in an implicit manner, thus possessing more powerful representation capacity than mesh-based templates. To the best of our knowledge, our work is the first one to learn implicit function-based templates for a collection of shapes.

\section{Overview}
\label{sec:overview}
Our Deep Implicit Template representation is designed on the basis of DeepSDF \cite{park2019deepsdf}, which is a popular DIF-based 3D shape representation. In this section, we first review DeepSDF for clarity and then describe the overall framework of our approach.

\subsection{Review of DeepSDF}
\label{sec:overview:deepsdf}
The DeepSDF representation defines a surface as the level set of a signed distance field (SDF) $\mathcal{F}$, e.g. $\mathcal{F}(\bm{p}) = 0$, where $\bm{p}\in\mathbb{R}^3$ denotes a 3D point and $\mathcal{F}\colon \mathbb{R}^3 \mapsto \mathbb{R}$ is a function approximated using a deep neural network. In practice, in order to represent multiple object instances using one neural network, the function $\mathcal{F}$ also takes a condition variable $\bm{c}$ as input and thus can be written as: 
\begin{equation}
\label{eqn:deepsdf}
    \mathcal{F}(\bm{p}, \bm{c}) = s\colon \bm{p}\in\mathbb{R}^3, \bm{c}\in\mathcal{X}, s\in\mathbb{R}
\end{equation}
where $\bm{c}\in\mathcal{X}$ is the condition variable that encodes the shape of a specific object and can be custom designed in accordance of applications \cite{chen2019implicit,park2019deepsdf,Mescheder2019OccupancyNetwork,pifuSHNMKL19}. With this SDF represented by $\mathcal{F}$, the object surface can be extracted using Marching Cube \cite{marching_cubes}.  In DeepSDF \cite{park2019deepsdf}, the condition variable $\bm{c}$ is a high-dimensional latent code and each shape instance has a unique code. All latent codes are firstly initialized with Gaussian noise and then optimized in parallel with network training. 

\begin{figure*}
	\centering
	\includegraphics[width=0.99\linewidth]{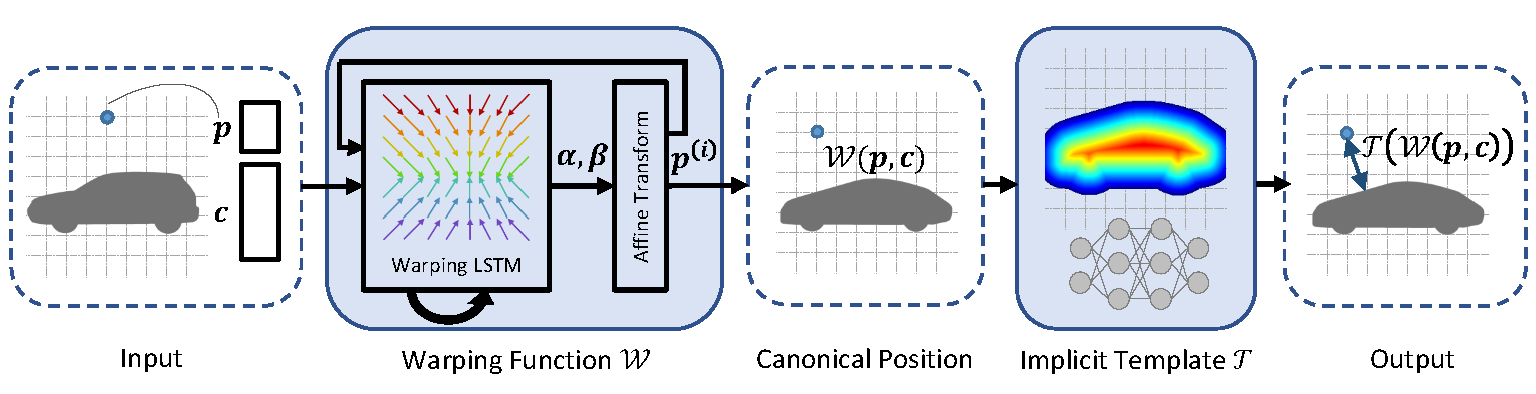}
	\caption{Method overview. Our method decomposes the DIF representation into a warping function and an implicit template. The warping function transforms point samples of shape $\bm{c}$ to their canonical positions, which are then mapped to SDF values by the implicit template. }
	\label{fig:arch}
\end{figure*}

\subsection{Deep Implicit Templates}
\label{sec:overview:overview}
In DeepSDF, the shape variance is directly represented by the changes of SDFs themselves. Different from this formulation, we think that, given a category of shapes represented by SDFs, their shape variance can be reflected by the differences of these SDFs relative to a template SDF that captures their common structure. 
This key idea leads to our formulation of \textit{Deep Implicit Templates}, which decomposes the conditional signed distance function $\mathcal{F}$ into $\mathcal{F}=\mathcal{T} \circ \mathcal{W}$, i.e.,  
\begin{equation}
\label{eqn:deep_implicit_template}
    \mathcal{F}(\bm{p}, \bm{c}) = \mathcal{T}(\mathcal{W}(\bm{p}, \bm{c})) 
\end{equation}
where $\mathcal{W}: \mathbb{R}^3\times \mathcal{X} \mapsto \mathbb{R}^3$  maps the coordinate of $\bm{p}$ to a new 3D coordinate, while $\mathcal{T}: \mathbb{R}^3 \mapsto \mathbb{R}$ outputs the signed distance value at this new 3D coordinate. 

Intuitively, $\mathcal{W}$ is a conditional spatial warping function that warps the input points according to the latent code $\bm{c}$, while the function $\mathcal{T}$ itself is an implicit function representing a common SDF which is irrelevant to $\bm{c}$. 
Therefore, for a set of objects, the shape represented by $\mathcal{T}(\cdot)$ can be regarded as their common template SDF; in the following context we call it an \textit{implicit template}. To query the signed distance at $\bm{p}$ for a specific object defined by $\bm{c}$, the spatial warping function $\mathcal{W}$ first transforms $\bm{p}$ to its \textit{canonical position} in the implicit template, followed by $\mathcal{T}$ querying its signed distance. 
In other words, the implicit template is warped according to the latent codes to model different SDFs.
\footnote{ In the strict sense of the term, a warping of an SDF is not an SDF in general. However, we adopt two constraints to make sure that the warped SDF can still approximate the target SDF: (1) we use truncated SDF that only varies in a small band near the surface, and (2) we normalize all the meshes into the same scale. Therefore, we loosely use the term "SDF" in this paper and regard a warping of an SDF as another SDF. 
}


Compared to the original formulation in Eqn.(\ref{eqn:deepsdf}), the main advantage of the decomposition in Eqn.(\ref{eqn:deep_implicit_template}) is that it naturally induces correspondences between the implicit template and object instances, and accordingly, correspondences across different object instances. As shown in Sec.\ref{sec:extension}, this feature offers more possibility for applying deep implicit functions in many applications. 

However, implementing and training the decomposed network is not straight-forward. 
Specifically, without proper design and regularization, the network tends to overfit to a complicated transformer with an over-simplified implicit template, which further result in inaccurate correspondences.
Our goal is to learn an optimal template that can represent the common structure for a set of objects, together with a spatial transformer that establishes accurate dense correspondences between the template and the object instances. Moreover, the learned Deep Implicit Templates should also preserve the representation power and the generalization capacity of DeepSDF, and thus support mesh interpolation and shape completion. In Sec.\ref{sec:method} we will discuss how we achieve these goals.


\section{Methodology}
\label{sec:method}
\subsection{Network Architecture}

Similar to DeepSDF \cite{park2019deepsdf}, we implement our implicit template, $\mathcal{T}$ in Eqn.(\ref{eqn:deep_implicit_template}), using a fully-connected network. For the spatial warping function $\mathcal{W}$, we empirically found that an MLP implementation leads to unsatisfactory results (Sec.\ref{sec:experiments:ablation}). 
To deal with this challenge, we introduce a Spatial Warping LSTM, which decomposes the spatial transformation for a point $\bm{p}$ into multi-step point-wise transformations:
\begin{equation}
\label{eqn:lstm}
\begin{aligned}
    (\bm{\alpha}^{(i)}, \bm{\beta}^{(i)}, \bm{\phi}^{(i)}, \bm{\psi}^{(i)}) &= \\
    \rm{LSTMCell}&(\bm{c}, \bm{p}^{(i-1)}, \bm{\phi}^{(i-1)}, \bm{\psi}^{(i-1)}), 
\end{aligned}
\end{equation}
where $\bm{\phi}$ and $\bm{\psi}$ are the output and cell states, $\bm{\alpha}$ and $ \bm{\beta}$ are the transformation parameters, and the superscript $(i)$ means the results of the $i$-th step. The position of $\bm{p}$ is updated as:
\begin{equation}
\label{eqn:update}
    \bm{p}^{(i)} = \bm{p}^{(i-1)} + (\bm{\alpha}^{(i)}\odot\bm{p}^{(i-1)} + \bm{\beta}^{(i)}). 
\end{equation}
where $\odot$ means element-wise product and $\bm{p}^{(i)}=\bm{p}$. We iterate this process in Eqn.(\ref{eqn:lstm}-\ref{eqn:update}) for $S$ steps (in all experiments we set $S=8$), and the point coordinate at the final step yields the output of the warping function $\mathcal{W}$:
\begin{equation}
\label{eqn:final_output_of_warping_function}
    \mathcal{W}(\bm{p}, \bm{c}) =  \bm{p}^{(S)}
\end{equation}
This multi-step formulation takes inspiration from the iterative error feedback (IEF) loop \cite{IEF}, where progressive changes are made recurrently to the current estimate. The difference is that our LSTM implementation allows us to aggregate information from multiple previous steps while IEF makes independent estimations in each step. The network architecture is illustrated in Fig.\ref{fig:arch}.


\begin{figure*}
	\centering
	\includegraphics[width=0.95\linewidth]{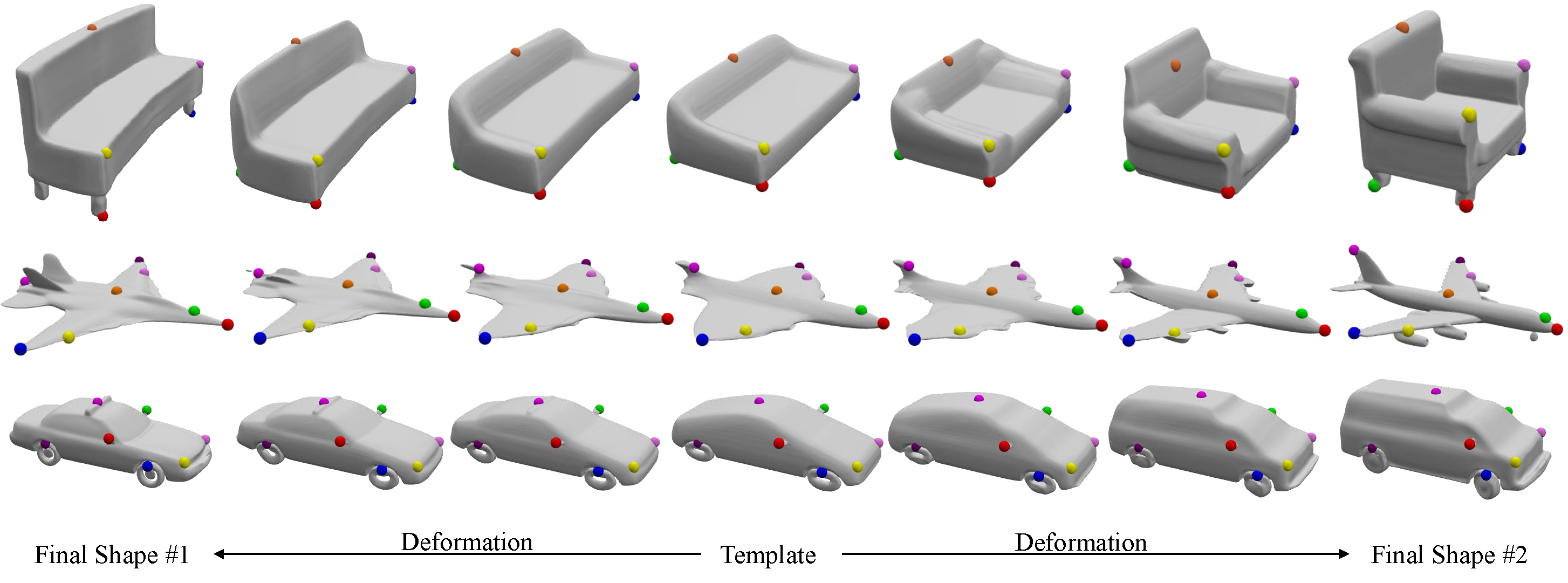}
	\caption{Demonstration of our representation. We manually select several points on the learned templates (middle) and track their traces when the templates deforms to form different shapes (leftmost and rightmost). Note that the deformation is defined and performed in an implicit manner; here we visualize the intermediate results by interpolating the warping fields. The selected points and their movements are rendered as small colored balls.  }
	\label{fig:deform}
\end{figure*}

\subsection{Network Training}
The training loss for our network is composed of two components, a reconstruction loss and a regularization loss:
\begin{equation}
    \mathcal{L} = \mathcal{L}_{rec} + \mathcal{L}_{reg}
\end{equation}
Below we will discuss them in details. 

\subsubsection{Progressive Reconstruction Loss}
As we decompose the spatial warping field into multiple steps of point-wise transformation using our Spatial Warping LSTM, 
we expect that the network learns a progressive warping function, which starts from obtaining smooth shape approximations and then gradually strives for more local details. 
To this end, we take inspiration from the shape curriculum in  \cite{duan2020cdeepsdf} and impose a \emph{progressive reconstruction loss} upon the outputs of our network for different numbers of warping steps, aiming that the network recovers more geometric details when taking more transformation steps. 
Mathematically, the loss term for the outputs with $s$ warping steps is defined as:
\begin{equation}
\label{eqn:data_loss_one_step}
    \mathcal{L}_{rec}^{(s)} = \sum_{k=1}^{K} \sum_{i=1}^{N} L_{\epsilon_s, \lambda_s}\left(\mathcal{T}\left(\bm{p}^{(s)}\right), v_{k,i}\right)
\end{equation}
where $\bm{p}^{(s)}$ is defined as in Eqn.\ref{eqn:update}, $v_{k,i}$ the ground-truth SDF value of $\bm{p}_i$ for the $k$-th shape, $N$ the number of SDF samples for one shape and $K$ the number of shapes. $L_{\epsilon, \lambda}(\cdot, \cdot)$ is a curriculum training loss with $\epsilon$ and $\lambda$ controlling its smoothness level and hard example weights; please refer to  \cite{duan2020cdeepsdf} for detail definition. In Tab.\ref{tab:data_loss_details} we present the parameter settings for different warping steps.  
Our progressive reconstruction loss is the sum of all levels of $\mathcal{L}_{rec}^{(s)}$:
\begin{equation}
\label{eqn:data_loss_total}
\mathcal{L}_{rec} = \sum_{s\in\{2, 4, 6, 8\}} \mathcal{L}_{rec}^{(s)} 
\end{equation}


\begin{table}
\centering
    \footnotesize
    \begin{tabular}{ccccccccc}
    \toprule   
        $s$ & 1 & 2 & 3 & 4 & 5 & 6 & 7 & 8 \\
    \midrule
        $\epsilon_s$ & -& 0.025 &-& 0.01 &-& 0.0025 &-& 0 \\
        $\lambda_s$ &-& 0 &-& 0.1 &-& 0.2 &-& 0.5 \\
    \bottomrule
    \end{tabular}
    \vspace{2pt}
    \caption{Parameter settings of the progressive reconstruction loss for different warping steps. For efficiency, we only construct loss every other steps. }
    \label{tab:data_loss_details}
\end{table}


\subsubsection{Regularization Loss}
Ideally, the spatial warping function is supposed to establish plausible correspondences between the template and the object instances, while the implicit field template should capture the common structure for a set of objects. 
To achieve this goal, we introduce two regularization terms on point-wise and point-pair levels for the warping function. 

\textbf{Point-wise regularization. } We assume that all meshes are normalized to a unit sphere and aligned in a canonical pose. Therefore, we introduce a point-wise regularization loss that is used to constrain the position shifting of points after warping. It is defined as: 
\begin{equation}
    \mathcal{L}_{pw} = \sum_{k=1}^{K} \sum_{i=1}^{N} h\left( \lVert \mathcal{W}(\bm{p}_i, \bm{c}_k) - \bm{p}_i\rVert_2 \right), 
\end{equation}
where $h(\cdot)$ is the Huber kernel with its hyper-parameter $\delta_h$. 

\textbf{Point pair regularization. } Although spatial distortion is inevitable during template deformation, extreme distortions should be avoided. To this end, we introduce a novel regularization loss for the point pairs in each shape:
\begin{equation}
    \mathcal{L}_{pp} = \sum_{k=1}^{K} \sum_{i\neq j} \max\left(\frac{\lVert\Delta\bm{p}_i-\Delta\bm{p}_j\rVert_2}{\lVert \bm{p}_i-\bm{p}_j \rVert_2} - \epsilon, 0\right), 
\end{equation}
where $\Delta\bm{p} = \mathcal{W}(\bm{p}, \bm{c}) - \bm{p}$ is the position shift of $\bm{p}$ and $\epsilon$ is a parameter controlling the distortion tolerance. Intuitively, if $\epsilon=0$, $\mathcal{L}_{pp}$ is reduced to a strict smoothness constraint enforcing translation consistency on neighboring points. 
We set $\epsilon = 0.5$ in our experiments to construct a relaxed formulation of smoothness loss, which is found important to prevent shape structures from collapsing (Sec.\ref{sec:experiments:ablation}). 

Our final regularization loss is defined as: 
\begin{equation}
    \mathcal{L}_{reg} = \lambda_{pw}\mathcal{L}_{pw} + \lambda_{pp}\mathcal{L}_{pp} + \frac{1}{\sigma^2} \sum_{k=1}^{K} \lVert\bm{c}_k\rVert_2^2, 
\end{equation}
where the last term is the same magnitude constraint on the latent codes as in DeepSDF  \cite{park2019deepsdf}. 

\begin{figure*}
	\centering
	\includegraphics[width=0.95\linewidth]{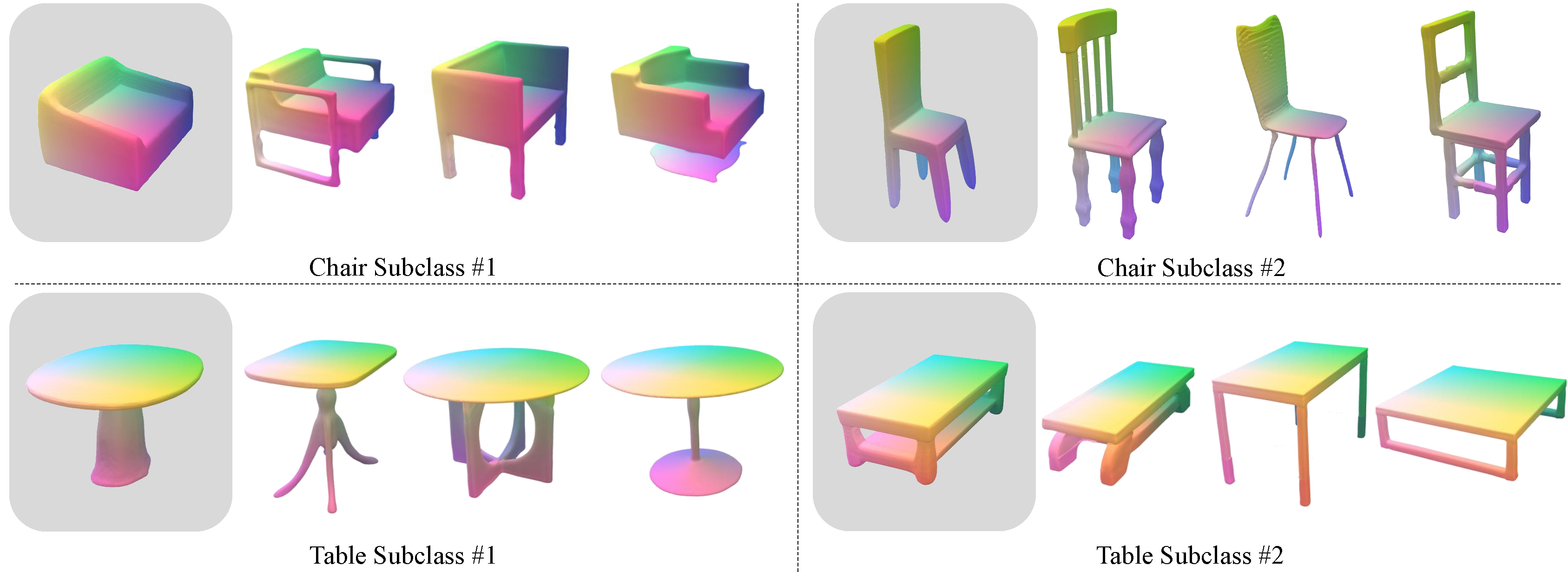}
	\caption{More demonstration of our representation. For objects that show very high structure varieties, like chairs and tables, we can train individual deep implicit templates for different subclass. The learned templates are rendered in gray background and correspondences are shown with the same color. }
	\label{fig:chair_table}
\end{figure*}

\begin{table*}
\centering
    \footnotesize
    \begin{threeparttable}
    \begin{tabular}{lccccccccccc}
        \toprule
         & \multicolumn{5}{c}{CD Mean} & \multicolumn{3}{c}{CD Median} & \multicolumn{3}{c}{EMD Mean}                  \\
        \cmidrule(r){2-6} \cmidrule(r){7-9} \cmidrule(r){10-12} 
        Model $\backslash$ Shape Class    & Airplanes(K) & Airplanes & Sofas & Cars & Chairs & Airplanes & Sofas & Chairs & Airplanes & Sofas & Chairs \\
        \midrule 
        AtlasNet-Sph \cite{AtlasNet2018}\tnote{$\ast$}      &	-&	    0.19&	0.45&	-&	    0.75&	0.079&	0.33&	0.51&	0.038&	0.050&	0.071 \\
        AtlasNet-25 \cite{AtlasNet2018}\tnote{$\ast$}       &	-&	    0.22&	0.41&	-&	    0.37&	0.065&	0.31&	0.28&	0.041&	0.063&	0.064 \\
        SIF \cite{SIF}\tnote{$\ast$}                        &	-&	    0.44&	0.80&	1.08&	1.54&	    -&	-&	-&	-&	-&	- \\
        DeepSDF \cite{park2019deepsdf}\tnote{$\dagger$}     &	0.05&	0.14&	0.12&	0.11&	0.24&	0.061&	0.08&	0.10&	0.035&	0.051&	0.055\\
        C-DeepSDF \cite{duan2020cdeepsdf}\tnote{$\dagger$}  &	0.03&	0.07&	0.11&	0.06&	\textbf{0.16}&	0.033&	0.07&	\textbf{0.06}&	\textbf{0.026}&	\textbf{0.044}&	\textbf{0.048}\\
        DualSDF \cite{zekun2020dualsdf}\tnote{$\dagger$}    &	0.19&	0.22&	   -&	-&	    0.45&	0.14&	   -&	0.21&	0.041&	-&	0.055\\
        \midrule 
        Ours (w/o Prog. Loss)\tnote{$\dagger$}              &	0.042&	0.104&	0.117&	0.093&	0.23&	0.040&	0.075&	0.113&	0.031&	0.047&	0.055\\
        Ours\tnote{$\dagger$}                               &   \textbf{0.025}&	\textbf{0.053}&	\textbf{0.093}&	\textbf{0.052}&	0.20&	\textbf{0.027}&	\textbf{0.061}&	0.071&	0.029&	0.046&	0.049\\
        \bottomrule
    \end{tabular}
    \end{threeparttable}
    \vspace{2pt}
    \caption{Reconstruction accuracy of different representations on known (K) shapes and unknown shapes for various object categories. Lower is better. (Mean and median Chamfer distance multiplied by $10^{3}$). We highlights methods based on auto-decoders ($\dagger$) from methods that use an encoder-decoder network architectures ($\ast$).  }
    \label{tab:recon_accuracy}
\end{table*}

\section{Experiments}
\label{sec:experiments}
\subsection{Experimental Setup}
We train Deep Implicit Templates on ShapeNet dataset  \cite{ShapeNet2015}, following  \cite{park2019deepsdf} for data pre-processing. 
In Sec.\ref{sec:experiments:results}, we first show that our model is capable of representing shapes in high quality with dense correspondences. 
For comparison in Sec.\ref{sec:experiments:comparison}, we select several strong baselines that use different types of representations: DeepSDF (deep implicit functions)  \cite{park2019deepsdf}, SIF (structured implicit functions)  \cite{SIF} , AtlasNet (mesh parameterization) \cite{AtlasNet2018}, PointFlow (point clouds) \cite{pointflow} and DualSDF (two-level implicit functions) \cite{zekun2020dualsdf}. We mainly evaluate them and our method in terms of reconstruction, correspondence and interpolation. 
Finally we evaluate our technical contributions in Sec.\ref{sec:experiments:ablation}. 
More results, experiments and details are presented in the supplemental materials.

\subsection{Results}
\label{sec:experiments:results}

We demonstrate some results of our approach in Fig.\ref{fig:teaser}, Fig.\ref{fig:deform} and Fig.\ref{fig:chair_table}.  
In Fig.\ref{fig:teaser}, we present the learned template shapes as well as the dense correspondences between the templates and object instances. 
The results show that our method can learn to  abstract a template that captures the common structure for a collection of shapes, and also establish plausible dense correspondences that indicates the semantic relationship across different shapes. The results also show that our representations can deal with large deformations and describe objects with completely different structures. 
In Fig.\ref{fig:deform}, we provide a more clear landscape on how the templates deforms to describe different objects. 
Fig.\ref{fig:chair_table} further presents the representation power of our method as well as the accurate correspondences established by our method. 

\subsection{Comparison}
\label{sec:experiments:comparison}

\noindent\textbf{Reconstruction.}  
We report the reconstruction results for known and unknown shapes (i.e., shapes belonging to the train and test sets) in Tab.\ref{tab:recon_accuracy}. We use two metrics for accuracy measurement, i.e., Chamfer Distance (CD) and Earth Mover Distance (EMD). 
As the numeric results show, our method is able to achieve comparable reconstruction performance when compared to state-of-the-art methods. Furthermore, we observe that our method is able to produce slightly better results for object classes that have a strong template prior (e.g., cars, airplanes and sofas), but degenerates for objects that vary enormously in shape structures (e.g., chairs).

\begin{figure}
	\centering
	\includegraphics[width=1.0\linewidth]{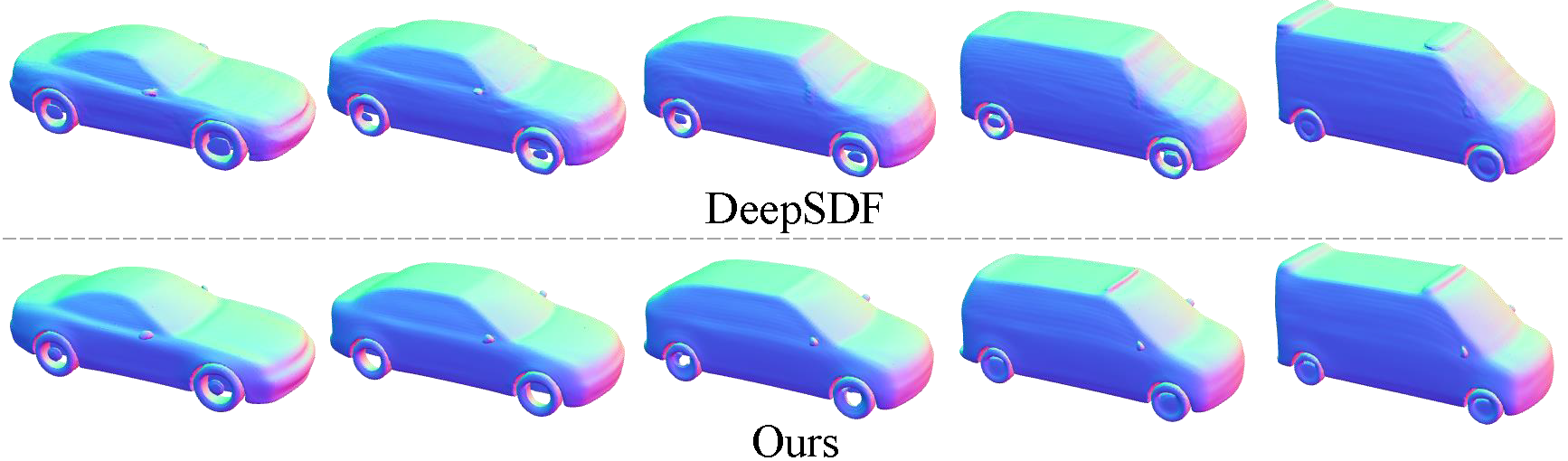}
	\caption{Shape interpolation results of DeepSDF and our method. Given the latent code of different meshes (leftmost and rightmost), we interpolate the latent codes linearly and generate the corresponding meshes (2nd - 4th colume). Models are rendered using their surface normals.}
	\label{fig:interp_comp}
\end{figure}

\noindent\textbf{Interpolation.} 
Similar to DeepSDF, our learned shape embedding is continuous and supports shape interpolation in the latent space (Fig.\ref{fig:interp_comp}). Note that unlike DeepSDF that directly interpolates signed distance fields, our representation actually \textit{interpolates the spatial warping fields} because the template is independent from the latent code. The results in Fig.\ref{fig:interp_comp} (bottom row) show that the generative capability of DeepSDF is well preserved in our representation.

\begin{figure}
    \centering
    \includegraphics[width=0.95\linewidth]{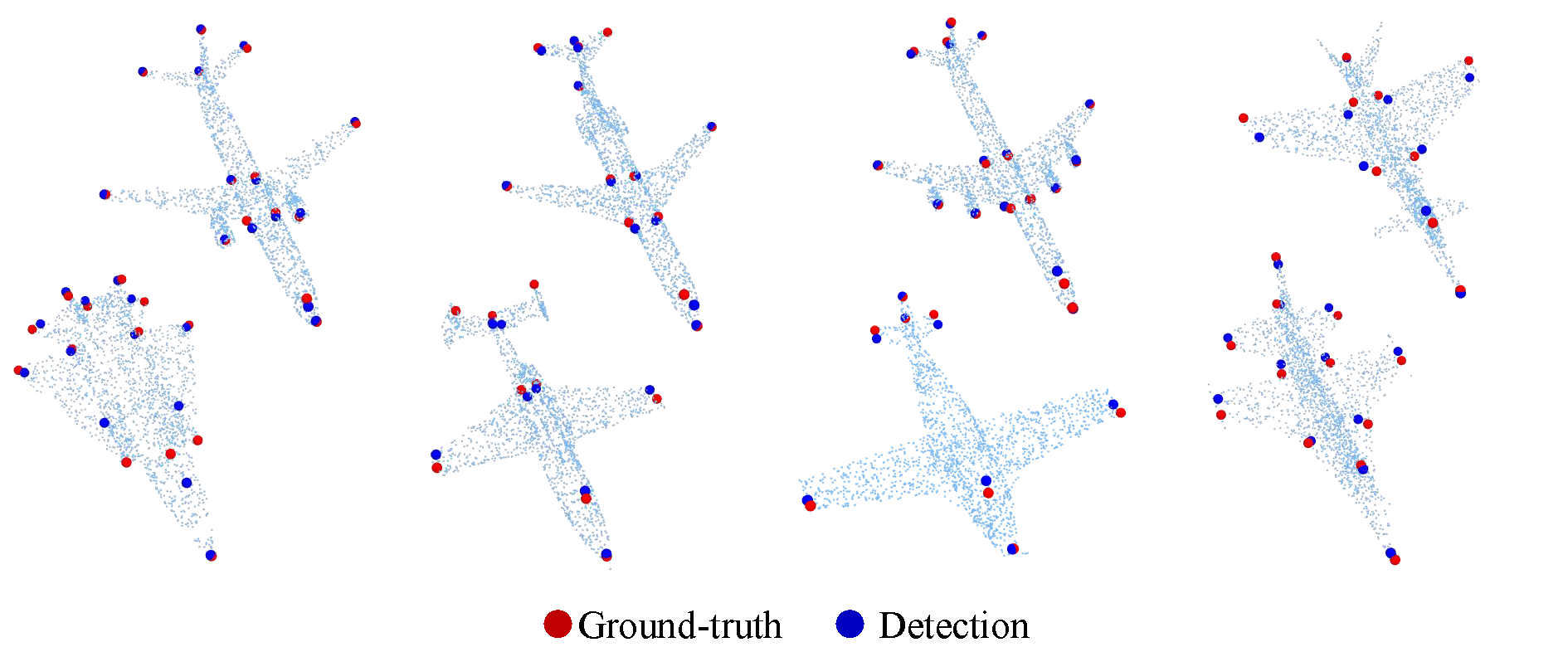}
    \caption{Keypoint detection results by our method. Note that although our network is not trained for keypoint detection, it can still output accurate detection.  }
    \label{fig:keypoint_detection}
\end{figure}

\noindent\textbf{Correspondences.} 
With the dense correspondence provided by our method, one can perform keypoint detection on point clouds by transferring the keypoint labels in training shapes to test ones. Therefore, we use this as a \emph{surrogate} to evaluate the accuracy of correspondences, as there is no large-scale dense correspondence annotation available.  We collect keypoint annotations from KeypointNet \cite{KeypointNet2020} and use the percentage of correct keypoints (PCK) \cite{SyncSpecCNN2017} as the metric for our experiments. Tab.\ref{tab:pck} presents the PCK scores under different error distance threshold, showing that our method outperforms other representations. In Fig.\ref{fig:keypoint_detection},  we can see that, although our network has no access to the keypoint annotations or correspondence annotations during training, it still establishes accurate corresponding relationship across different objects in the same category. 



\begin{table}
    \centering
    \footnotesize
    \begin{tabular}{lcc}
        \toprule   
            Model $\backslash$ Shape Class & Planes & Cars \\
        \midrule
            SIF \cite{SIF}                      & 0.267 / 0.446     & 0.262 / 0.402 \\
            PointFlow \cite{pointflow}          & 0.286 / 0.573     & 0.083 / 0.326 \\
            AltasNet-25 \cite{AtlasNet2018}     & \textbf{0.411} / 0.628     & 0.206 / 0.361 \\
        \midrule
            Ours (w/o point-pair reg)           & 0.306 / 0.474     & 0.305 / 0.480\\
            Ours                                & 0.365 / \textbf{0.652}     & \textbf{0.345} / \textbf{0.530} \\
        \bottomrule
    \end{tabular}
    \vspace{2pt}
    \caption{Correspondence accuracy comparison. We report the PCK scores with threshold of $0.01$ / $0.02$ for keypoint detection of different methods. Higher is better. }
    \label{tab:pck}
\end{table}

\subsection{Ablation Study}
\label{sec:experiments:ablation}
\begin{figure}
	\centering
	\includegraphics[width=1.0\linewidth]{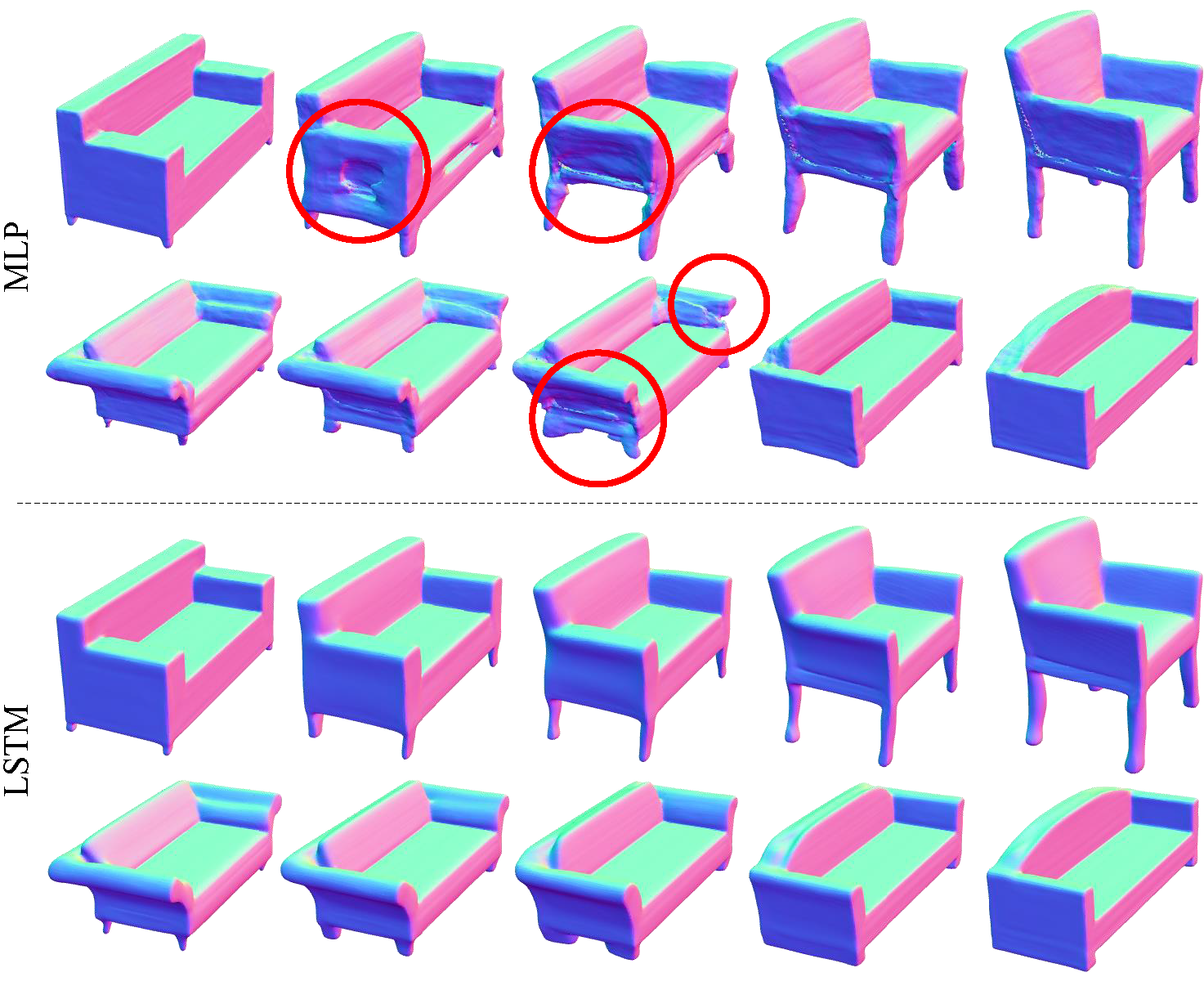}
	\caption{Interpolation capacity comparison of different architectures for the spatial warping function $\mathcal{W}$. Models are rendered using their surface normals.}
	\label{fig:interp_eval}
\end{figure}

\noindent\textbf{Spatial Warping LSTM.} 
We evaluate our choice of spatial warping LSTM by replacing it with an MLP-based implementation. The numbers of the MLP neurons are $(259, 512, 512, 512, 512, 512, 6)$. The comparison on interpolation capacity of different network architectures is presented in Fig.\ref{fig:interp_eval}.  We empirically find that the MLP implementation is prone to over-fitting and cannot generalize well for latent space interpolation. We think that it is because the large non-uniform deformations between the template and object instances are more easily to learn in a gradual manner, but much harder when an MLP try to learn the transformation in a single step. 

\noindent\textbf{Progressive Reconstruction Loss.}
We evaluate the effect of our progressive reconstruction loss in terms of reconstruction accuracy in Tab.\ref{tab:recon_accuracy} (last two rows). 
To construct the evaluate baseline, we replace the progressive reconstruction loss with the original $\ell_1$ loss in \cite{park2019deepsdf} and only apply supervision on the final output in Eqn.\ref{eqn:final_output_of_warping_function}.  The numeric results show that 
Note that we use the same settings of shape curriculum as \cite{duan2020cdeepsdf}; interestingly, we find that our method can even outperform \cite{duan2020cdeepsdf} on some object categories although we add additional regularization that may adversely impact the reconstruction. 
We speculate that this is because we fix the network structure during training and apply multi-level supervision simultaneously, while \cite{duan2020cdeepsdf} gradually increases the network depth and changes loss parameters as training proceeds, which may lead to some extent of instability for network training. 

\begin{figure}
	\centering
	\includegraphics[width=1.0\linewidth]{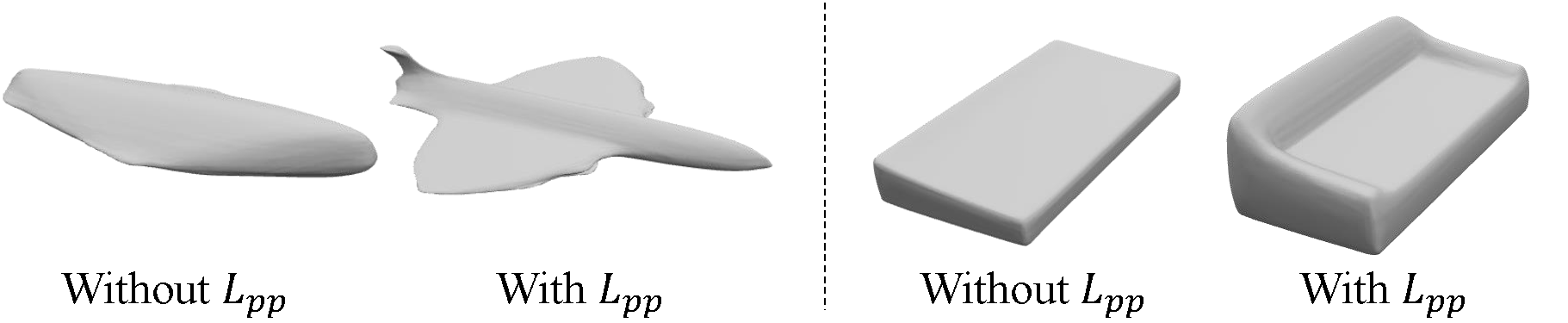}
	\caption{Evaluation of our point-pair regularization. Without the point-pair regularization, the network tends to learn over-simplified templates for planes and sofas. }
	\label{fig:pointpair_reg_eval}
\end{figure}

\noindent\textbf{Point-pair Regularization.}  
To evaluate our point-pair regularization loss, we train a baseline network without it. As shown in Fig.\ref{fig:pointpair_reg_eval}, without the point-pair regularization, the networks tends to learn over-simplified templates. Although this phenomenon has no impact on the reconstruct accuracy, it leads to inaccurate correspondences since the detailed structures like plane wings and sofa backrests collapse into tiny regions on the templates, as proven in Tab.\ref{tab:pck} (last two rows). Therefore, the proposed two-level regularization is essential in our representation of deep implicit templates. 

\section{Extensions and Applications}
\label{sec:extension}
\subsection{Extensions}
To prove the flexibility and the potential of our Deep Implicit Templates, we show how our method can be easily extended when more constraints are available.

\begin{figure}
	\centering
	\includegraphics[width=0.9\linewidth]{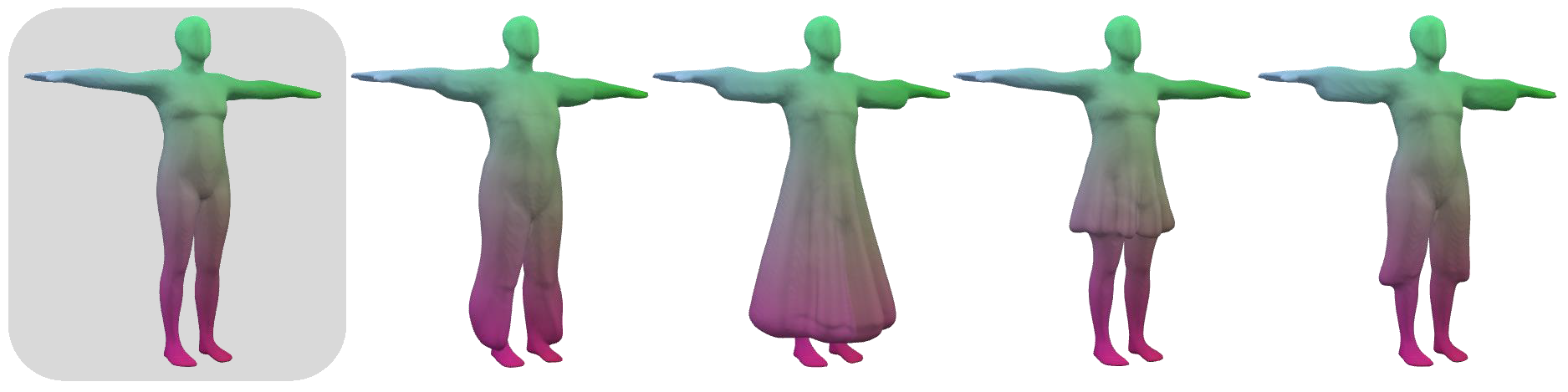}
	\caption{Results on clothed humans. With SMPL model as the manually specified template (leftmost), our method can learn to generate human models in various clothes.  }
	\label{fig:human_modeling}
\end{figure}

\begin{figure}
	\centering
	\includegraphics[width=1.0\linewidth]{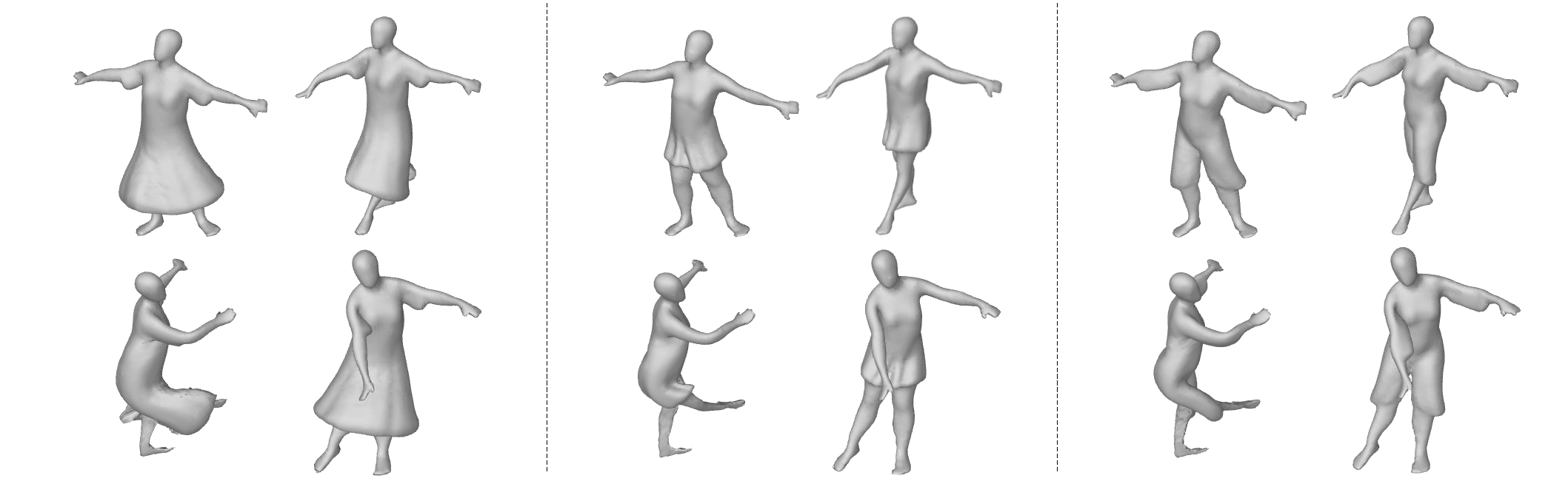}
	\caption{Human model animation. The dense correspondences between the SMPL template and the clothed human models allows automatic calculation of skining weights.  }
	\label{fig:human_skinning}
\end{figure}

\noindent\textbf{User-defined templates.} For example, we can replace the learned template in our method with a manually-specified one. 
In this case, we just need to construct an additional loss to train the template implicit function $\mathcal{T}$. Specifically, the loss is defined as: 
\begin{equation}
    \mathcal{L}_{temp} = \sum_{i=1}^{N} |\mathcal{T}(\bm{p}_i) - s_{i}|,
\end{equation}
where $(\bm{p}_i, s_i)$ are the SDF training pairs extracted from the specified template model. In Fig.\ref{fig:human_modeling}, we demonstrate an example where we specify SMPL model \cite{SMPL2015} as the template and train our network to model various clothed humans. With the correspondences we can transfer skinning weights from SMPL model to clothed humans, allowing animation for various clothed human models as in Fig.\ref{fig:human_skinning}.

\begin{figure}
\centering
\footnotesize

\begin{minipage}{0.9\linewidth}
    \centerline{\includegraphics[width=1.0\linewidth]{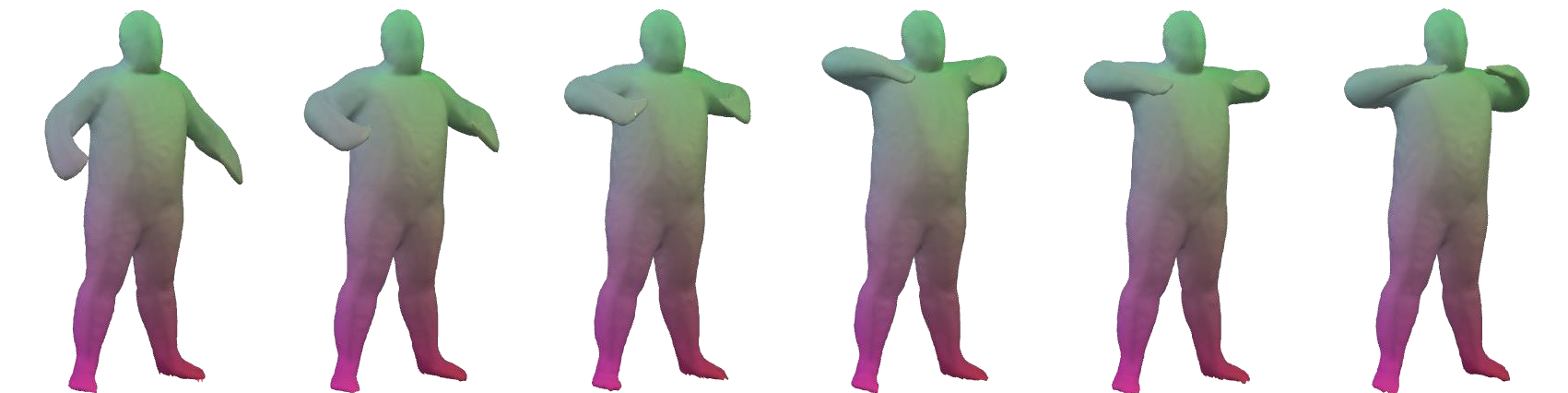}}
    \centerline{(a) Reconstruction results. }
\end{minipage}
\vspace{4pt}

\begin{minipage}{1.0\linewidth}
    \centering
    \footnotesize
    \begin{tabular}{ccc}
    \toprule   
          & Without $\mathcal{L}_{corr}$ & With $\mathcal{L}_{corr}$ \\
    \midrule
        Error & 0.337 & 0.085\\
    \bottomrule
    \end{tabular}
    \centerline{\quad }
    \centerline{(b) Correpsondence error \cite{OccupancyFlow}. Low is better. }
\end{minipage}
\caption{Results with the correspondence loss. If correspondence annotations are available, our method can learn to reconstruct human models with more accurate correspondences.  }
\label{fig:corre_loss}
\end{figure}

\noindent\textbf{Correspondence annotations.} Furthermore, although our method is designed to learn dense correspondence without any supervision, our method can incorporate with sparse/dense correspondence annotations when provided. To this end, we just need to introduce an additional loss to enforce the known correspondences:
\begin{equation}
    \mathcal{L}_{corr} =  \sum_{\substack{1\leq k,l\leq K\\k\neq l}} \sum_{(\bm{p}, \bm{q})\in \mathrm{C}_{k,l} } \lVert\mathcal{W}(\bm{p}, \bm{c}_k) - \mathcal{W}(\bm{q}, \bm{c}_l)\rVert_2^2,
\end{equation}
where $\mathrm{C}_{k,l}$ is the correspondence set of the $k$-th and $l$-th objects. Fig.\ref{fig:corre_loss} demonstrate the qualitative and quantitative results of using the correspondence loss when training our network on D-FAUST dataset \cite{dfaust}.

\subsection{Applications}

\begin{figure}
    \centering
    \footnotesize
    \subfigure[Texture transfer. ]{
        \begin{minipage}[t]{1.0\linewidth}
        \centering
        \includegraphics[width=0.9\linewidth]{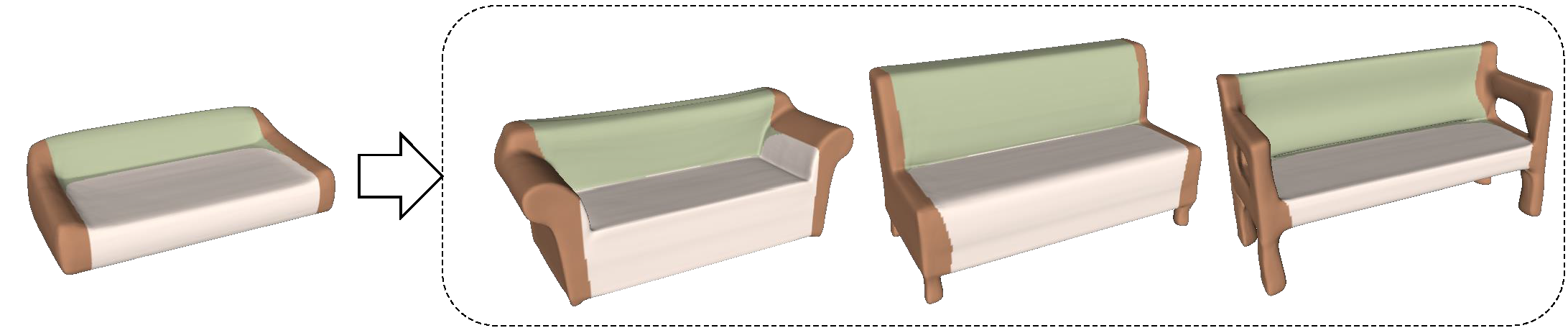}
        \end{minipage}%
    }%

    \subfigure[Shape manipulation. ]{
        \begin{minipage}[t]{1.0\linewidth}
        \centering
        \includegraphics[width=0.9\linewidth]{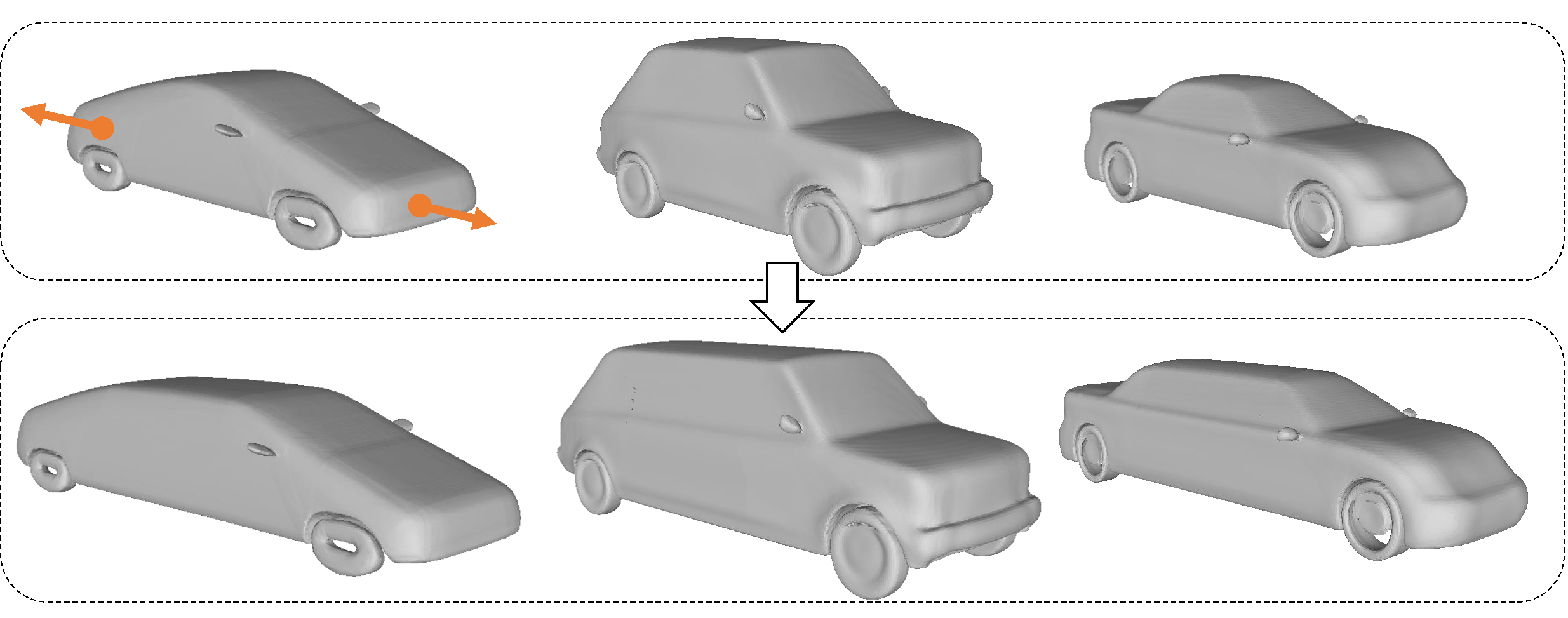}
        \end{minipage}%
    }%
    \caption{Our representation can be used in various applications. Please zoom in for better views.  }
    \label{fig:app}
\end{figure}




To show the effectiveness and practicability of our method, we investigate how Deep Implicit Templates can be used in applications. With the dense correspondences between the template and the object instances, we can transfer mesh attributes, such as texture (Fig.\ref{fig:app}(a)), or mesh stretching operation (Fig.\ref{fig:app}(b)) to multiple objects. 
Our method can also be used in the applications that have been demonstrated in DeepSDF \cite{park2019deepsdf} such as shape completion. 
\section{Conclusion}
\label{sec:conclusion}
We have presented Deep Implicit Templates, a new 3D shape representation that factors out implicit templates from deep implicit functions. 
To the best of our knowledge, this is the first method that explicitly interprets the latent space for a class of objects as a meaningful pair: an implicit template with its conditional deformations. 
Several technical contributions on network architectures and training losses are proposed to not only recover accurate dense correspondences without losing the properties of deep implicit functions, but also learn a plausible template capturing common geometry structures. 
The experiments further demonstrate some promising applications of Deep Implicit Templates. 
To conclude, we have demonstrated that a semantic interpretation of deep implicit functions leads to a more powerful representation, which we believe is an inspiring direction for future research. 


\noindent\textbf{Acknowledgement } This paper is supported by the NSFC No.61827805.

{\small
\bibliographystyle{ieee_fullname}
\bibliography{egbib}
}

\end{document}